\documentclass[11pt,a4paper]{article}
\usepackage[hyperref]{emnlp2018}
\usepackage{times}
\usepackage{latexsym}
\usepackage{times}
\usepackage{latexsym}
\usepackage{amsmath}
\usepackage{graphicx}
\usepackage{algorithm,algorithmic}
\usepackage{url}
\usepackage{color}
\usepackage{booktabs}
\usepackage{url}
\aclfinalcopy 

\title{Modeling Topical Coherence in Discourse without Supervision}

\author{Disha Shrivastava \thanks{   *Work done as part of IBM Research, Bangalore} \\
  MILA, Universit\'e de Montr\'eal\\
   Montreal, Canada \\
  {\tt dishu.905@gmail.com} \\\And
  Abhijit Mishra \\
  IBM Research\\
   Bangalore, India \\
  {\tt abhijimi@in.ibm.com} \\\And
  Karthik Sankaranarayanan \\
  IBM Research\\
   Bangalore, India \\
  {\tt kartsank@in.ibm.com} \\}

\date{}

\begin{document}
\maketitle
\begin{abstract}
Coherence of text is an important attribute to be measured for both manually and automatically generated discourse; but well-defined quantitative metrics for it are still elusive. In this paper, we present a metric for scoring topical coherence of an input paragraph on a real-valued scale by analyzing its underlying topical structure. We first extract all possible topics that the sentences of a paragraph of text are related to. Coherence of this text is then measured by computing: (a) the degree of uncertainty of the topics with respect to the paragraph, and (b) the relatedness between these topics. All components of our modular framework rely only on unlabeled data and WordNet, thus making it completely unsupervised, which is an important feature for general-purpose usage of any metric. Experiments are conducted on two datasets - a publicly available dataset for essay grading (representing human discourse), and a synthetic dataset constructed by mixing content from multiple paragraphs covering diverse topics. Our evaluation shows that the measured coherence scores are positively correlated with the ground truth for both the datasets. Further validation to our coherence scores is provided by conducting human evaluation on the synthetic data, showing a significant agreement of 79.3\%. 


\end{abstract}

\section{Introduction}
\label{sec:into}
Discourse coherence measurement has been an important task for evaluating human generated text \cite{attali2004automated,crossley2011text,taghipour2016neural} and text output produced by natural language generation (NLG) systems such as summarizers, descriptive question answering systems and automatic creative content generators. Measuring coherence is an essential need for evaluation of NLG systems that are currently impeded by lack of robust quality-estimation metrics \cite{belz2006shared,nenkova2007pyramid}. A robust evaluation metric may not only evaluate and interpret NLG systems better, but can also contribute to better system designs (such as using the metric as a loss or reward function in risk minimization or reinforcement learning settings).

\begin{table}[t]
\centering
\footnotesize
\begin{tabular}{p{7cm}}
\toprule
\textbf{Example 1 (Coherent):} \emph{The most important part of an essay is the thesis statement. The thesis statement introduces the argument of the essay. It also helps to create a structure for the essay. Therefore, one should always begin with a thesis statement while writing an essay.}\\
\midrule
\textbf{Example 2 (Locally Incoherent):} \emph{It also helps to create a structure for the essay. The thesis statement introduces the argument of the essay. The most important part of an essay is the thesis statement.}\\
\midrule
\textbf{Example 3 (Topically Incoherent):} \emph{The most important part of an essay is the \textbf{thesis statement}. \textbf{Essays} can be written on various topics from \textbf{domains} such as politics, sports, current affairs etc. I like to write about Cricket because it is the most \textbf{popular team sport} played at international level.}\\
\bottomrule
\end{tabular}
\label{ref:example}
\caption{Example of topically coherent, locally incoherent, and topically incoherent  paragraphs}
\end{table}
We propose a metric for topical coherence (referred to as coherence, henceforth) in paragraphs. Our work is motivated from the fact that automatic NLG systems (such as abstractive summarizers) often produce text with topics that may be quite unrelated to each other. For example, in a recently proposed work \newcite{see-liu-manning:2017:Long}, the summary generated by \textsc{Pointer-Gen+Coverage} system (Figure 1 in \newcite{see-liu-manning:2017:Long}), has a considerable topic shift (from ``administration'' to ``winning the election''), a mistake that is more likely to be committed by machines than humans. 

Topical coherence differs from local coherence. 
To illustrate the difference, Table \ref{ref:example} shows examples of coherent, locally incoherent and topically incoherent paragraphs. Example $1$  is quite coherent as it revolves around one central topic \textit{i.e.,} \emph{importance of thesis statement}. Example $2$ is more incoherent as compared to Example $1$, due to the fact that sentences are not ordered naturally. However, it still talks about the same central topic as Example $1$. Example $3$ on the other hand, confusingly covers multiple topics such as \emph{importance of thesis statement}, \emph{possible essay domains}, and \emph{popular sports}, which are quite unrelated to each other and hence, is topically incoherent.
%
Our metric for topical coherence is designed to address two key aspects: (a) how effectively each sentence contributes to the topic(s) that the paragraph expresses, and (b) to what extent topics expressed by the paragraph are related to each other. These aspects are tackled by independent modules, making our metric \textbf{modular and interpretable}. Typically, gold labels for coherence scores are difficult to obtain, and therefore the \textbf{unsupervised nature of our framework along with its simplicity} makes it easier and convenient to employ. We perform experiments on different sets of a publicly available human graded essay dataset, and a synthetic dataset constructed by injecting incoherence into already available coherent paragraphs (possibly mimicking machine generated discourse). Evaluation results show a positive correlation of the measured coherence scores with the gold-standard scores, hence making our system acceptable for coherence based ranking of paragraphs. Additionally, human evaluation on a set of synthetic data essays shows a significant agreement (79.3\%), demonstrating the effectiveness of our proposed metric. Finally, a comparison with the supervised systems discussed in \cite{barzilay2008modeling} shows that our framework, even though unsupervised and less complex, can exhibit performance competitive to the supervised systems that require significant amount of training data and in some cases, deep linguistics meta information such as role labels, coreferences, dependency parses \textit{etc.}, to produce reasonable results.
\section{Methodology}
\label{sec:component}
\subsection{Formulation}
\label{sec:central}
Based on the motivation laid above, a \textit{prima-facie} formulation of coherence of a paragraph (denoted as $CS$) can be proposed assuming that $CS$ varies linearly and positively with: (a) to what degree of certainty topics expressed by the paragraph (denoted as $T$) are supported by its constituent sentences (denoted by variable $E(T)$), and (b) to what extent these topics are related to each other (denoted by variable $\lambda(T)$). Mathematically, 
\begin{equation}
CS = \kappa \times E(T) \times \lambda (T)
\label{eq:main_coh}
\end{equation}
For empirical purposes, $\kappa$, the constant term can be set to $1$.
The two components of the above formula require topics expressed by the paragraph to be extracted first. This process is explained below.
\subsection{Defining Topics}

The first step before extracting the topics (which are abstract ``concepts'') is to define a set of wide variety of generic topics which could be mapped to any given paragraph. This one-time process is carried out by performing word-clustering on a large scale mixed domain unlabeled data. The assumption here is that clusters extracted from large scale data represent generic topics in the universal space of English language. After defining generic topics, the next step is to infer topics discussed in a given input paragraph. We explain each of these steps in detail below. 

For word clustering, we consider embeddings learned on a large scale corpora \cite{pennington2014glove} and cluster them via k-means clustering. The assumption here is that words connected to a particular topic would exhibit strong syntagmatic and paradigmatic relations. Since word embeddings are good at capturing such relations, topically connected words may eventually lie in the same cluster in the vector space. Each cluster in this space hence represents a \textit{topic}. Topic models such as the ones based on Latent Dirichlet Allocation \cite{blei2003latent} are other alternatives for topic extraction. However, it is well known that the behavior of topic models changes drastically\footnote{basic LDA based experiments did not show good performance in our setting} based on the prior distribution, hyper-parameters and document processing applied \cite{chang2009reading}. Hence, we rather keep the topic extraction process simplistic by using word embeddings.  

\subsection{Extracting Relevant Topics}
\label{subsec:themes}
Each sentence in the input paragraph is POS tagged and only nouns are selected for topic extraction. The intuition behind choosing only nouns is that nouns are most representative of the topic in a given sentence. For each noun, we determine the topic-cluster to which it belongs. To avoid noise, for each sentence, we choose a dominant cluster by assigning cluster scores to all clusters identified within a given sentence (the reason we try to extract topics for each sentence separately is that sentences are the atomic units that are capable of discussing independent topics from others). The cluster topic score of a cluster is (a) directly proportional to the fraction of nouns within the sentence that belong to the cluster and the mean pairwise cosine similarity of nouns in the sentence, and (b) is inversely proportional to the diameter of the cluster. The diameter of the cluster is the maximum distance between any two points in the cluster. The dominant cluster for a sentence is the one which has the highest cluster topic score. After determining the dominant cluster, we take all points within the cluster and check for its existence in WordNet. This eliminates highly specific terms, jargon, named entities which could potentially add noise to the clusters. Therefore, each topic can now be represented as a subgraph of the WordNet where the nodes in the subgraph are bag-of-words representing a topic (topicBOW).  

Given the topics relevant to the paragraph, we now explain the two components $E(T)$ and $\lambda(T)$ of Equation \ref{eq:main_coh} in the following subsections. 

\begin{figure}[t]
\begin{center}
\includegraphics[width=7cm]{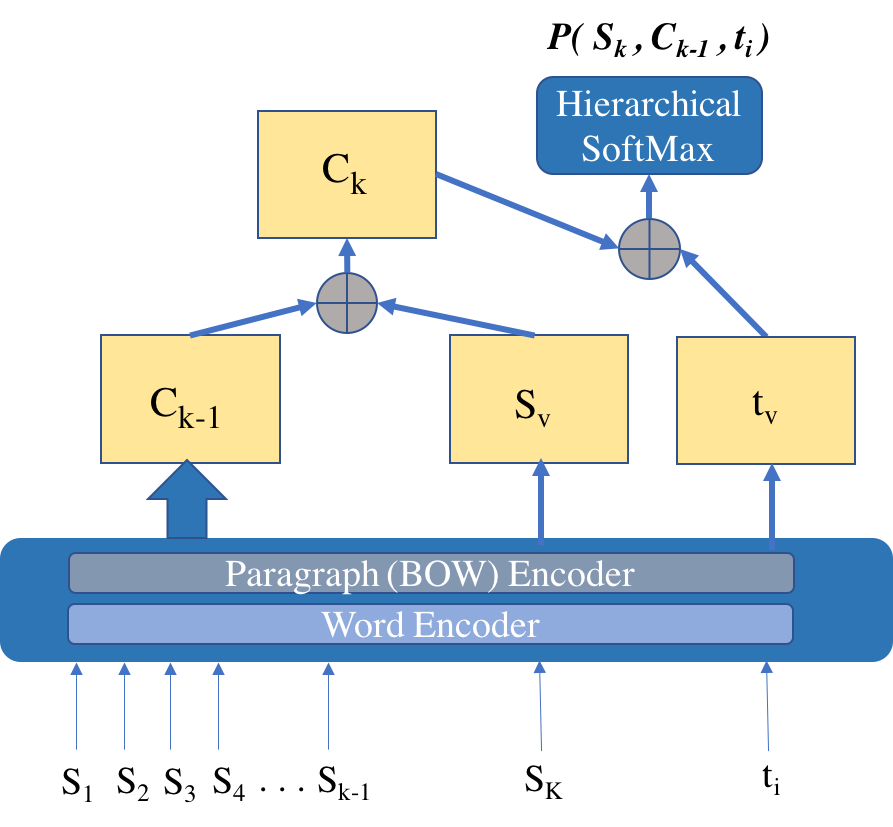}
\caption{Architecture for estimating $p(P,t_i)$}
\label{fig:sent_prob}
\end{center}
\end{figure}

\subsection{Component $E(T)$}
\label{subsec:entropy}
For an input paragraph $P$, with $M$ sentences ($s_1,s_2,...,s_M$), expressing a set of topics ($T=t_1,t_2,...,t_N$), the component $E(T)$ can be expressed as:
\begin{equation}
E(T) = -\sum_{i=1}^{N} p(t_i|P) log(p(t_i|P))
\label{eq:ET}
\end{equation}

where $p(t_i|P)=p(t_i|s_1,s_2,...,s_M)$ represents the probability of the topic $t_i$ conditioned over the paragraph. $E(T)$ corresponds to the \emph{conditional entropy} of topics here. Intuitively, if sentences in the paragraph are well distributed across all the topics emerging out of the paragraph, \textit{i.e.} each sentence in the paragraph is somewhat related (even though loosely) to all the topics, the paragraph may exhibit more coherence, with less topic-shift. On the other hand, if the paragraph can be divided in such a way that each segment of the paragraph is related to a unique topic, the paragraph will be less coherent. This is adequately captured by the conditional entropy formula which rewards when the distribution $p(T|P)$ is smooth and penalizes when it is sparse. Moreover, this formulation of $E(T)$ captures another essential aspect: if the number of topics in the paragraph become very large, it inherently pulls the conditional probabilities $p(t_i|P)$ down, making the distributions sparse\footnote{it is highly improbable to have sentences discussing a large number of topics} and thus, reducing the entropy. So, when number of topics grow, $E(T)$ is penalized more. 

The term $p(t_i|P)$ in Equation \ref{eq:ET} can be expanded further, with the help of Bayes' rule as follows:
\begin{equation}
p(t_i|P) = \frac{p(P|t_i)p(t_i)}{\sum_{j=1}^{N} p(P|t_j)p(t_j)}
\label{eq:Bayes}
\end{equation}

Furthermore, term $p(P|t_i)$ can be expanded using chain rule, as:
\begin{equation}
\begin{aligned}
&p(P|t_i) = p(s_1,s_2,...,s_M | t_i)\\
&= p(s_1|t_i) p(s_2|s_1,t_i)... p(s_M|s_1,s_2,...,s_{M-1},t_i)\\
&= \prod_{k=1}^{M}p(s_k|C_{k-1},t_i)
\end{aligned}
\label{eq:Chain}
\end{equation}


Here $C_{k-1}$ can be considered as the \textit{context} that appears before the the $k_{th}$ sentence.

Since, no topic in the paragraph can be given more importance over another, the probability term $p(t_i)$ can be uniformly distributed across $T$ \textit{i.e.,} $p(t_i) = \frac{1}{N}$. For estimating $p(s_k|C_{k-1},t_i)$ (without applying any simplifying assumptions), we follow the distributed bag-of-words (BOW) model by \newcite{le2014distributed}. The idea here is to train distributed bag-of-words model using a large number of sentences (covering good number of topics); and later on use the model for inferring $p(s_k|C_{k-1},t_i)$. The inference time snapshot of the model is given in Figure \ref{fig:sent_prob} and the probability estimation algorithm is given in Algorithm \ref{alg:estimate}. 

\algsetup{indent=2em}
\newcommand{\estimate}{\ensuremath{\mbox{\sc Probability Estimation }}}
\begin{algorithm}[t]
\caption{$\estimate$}
\label{alg:estimate}

\begin{algorithmic}[1]

\STATE $model$ $\gets$ TrainDistBOWModel
\STATE {\bf REM} TrainDistBOWModel is the process of training a distributed representation model for bag-of-words with unlabeled data.
\newline

\STATE $C_0$ $\gets$ $null$
\STATE $t_v$ $\gets$ InferBOWVector ($t_i$)
\STATE {\bf REM} InferBOWVector infers encoded vector for topic $t_i$ from the distributed representation model. 
\STATE $p(P|t_i) = 1$
\newline
\FOR {$s_k \in P={s_1,s_1,...,s_M}$}
	\STATE $s_v$ $\gets$ InferBOWVector ($s_k$)
	\STATE $C_{k} = C_{k-1} \oplus s_v$
    \STATE $C_{curr} = C_{k} \oplus t_v$
    \STATE $C_{prev} = C_{k-1} \oplus t_v$
    \STATE $p(s_k,C_{k-1},t_i)$ $\gets$ InferProb($C_{curr}$)
    \STATE {\bf REM} InferProb produces the joint probability score of a given bag-of-word distributed representation.
    \STATE $p(C_{k-1},t_i)$ $\gets$ InferProb($C_{prev}$)
    
    \STATE $p(s_k|C_{k-1},t_i) = \frac{p(s_k,C_{k-1},t_i)}{p(C_{k-1},t_i)}$
    \STATE $p(P|t_i) = p(P|t_i) \times p(s_k|C_{k-1},t_i)$
\ENDFOR
\renewcommand{\algorithmicreturn}{\textbf{Output:}}
\RETURN $p(P|t_i)$
\end{algorithmic}
\end{algorithm}

As discussed earlier in Section \ref{subsec:themes}, since topics are also treated as bag of words (topicBOW), it is easy to do conditional inference using a bag-of-words based distributed representation model. In such a setting, both sentences and topics are treated as bag-of-words, and hence, there is no special signal to be passed to the model with respect to the topic bag-of-word representations. We agree that bag-of-words based techniques can be agnostic to within-sentence sequentiality and natural order of sentences. However, since  modeling topical coherence involves inferring sentence-topic associations, and does not necessarily aim to model within-sentence properties; not preserving sentence order would not adversely affect our goal. We now describe the second component of Equation \ref{eq:main_coh}. 
\subsection{Component $\lambda(T)$}
\label{subsec:lambda}
The component $\lambda(T)$ aims to capture \emph{relatedness} between topics expressed by the paragraph. For two paragraphs with same number of topics, the coherence score should be more if \textbf{topics are strongly related with each other} as compared to the case where the correlation between topics of the paragraph is less. This helps us to refrain from penalizing the coherence scores even though the paragraph contains large number of topics if such topics are  strongly related. 
For example, a topical shift from the topic \emph{carnivore} to \emph{mammals} in general should be penalized less than that from \emph{carnivore} to \emph{electronics}.
The inter-relatedness between topics is captured by $\lambda(T)$, for which we rely on lexical knowledge networks such as WordNet \cite{fellbaum1998wordnet}, that preserve various forms of conceptual-semantic and lexical relations between words and are well-curated. Note that, we do not opt for simplistic topic-relatedness measures such as inter-cluster distance between topic-clusters, as such measures are significantly affected by the bias in the data used for clustering, and noise introduced by imperfect algorithm and parameter selection.
\begin{figure*}[t]
\begin{center}
\includegraphics[width=11.5cm]{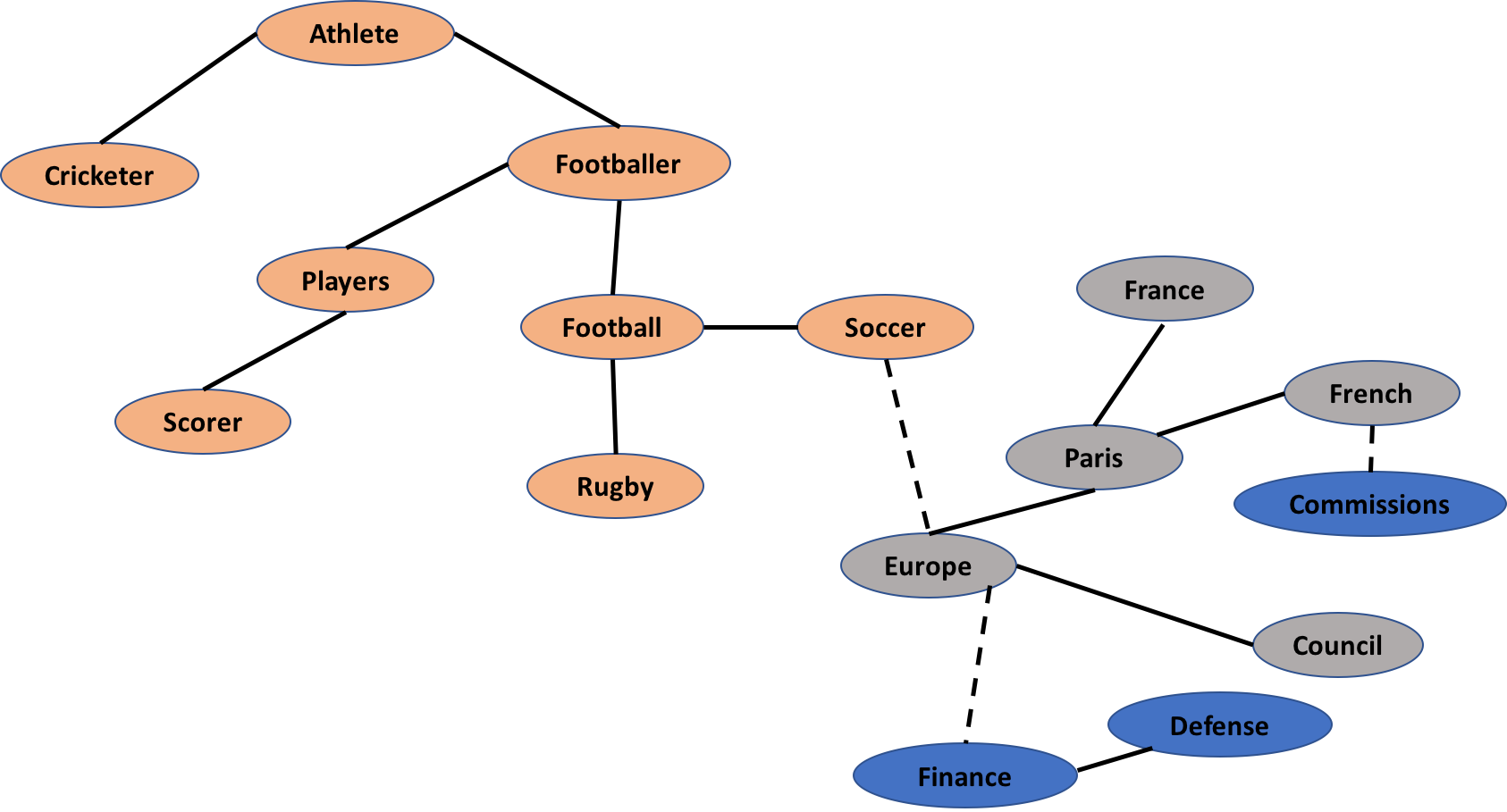}
\caption{ Sample WordNet subgraph extracted for calculating $\lambda(T)$. Dashes indicate indirect connections.}
\label{fig:wngraph}
\end{center}
\end{figure*}

Each word in the topic bag-of-words (for the whole paragraph) is mapped to a node in the WordNet (if its lemma exists in WordNet), according to its most frequent sense. The mapped nodes in the WordNet are then connected with each other through other intermediate nodes to form a subgraph. Figure \ref{fig:wngraph} illustrates a subgraph extracted for Example 2 discussed in Section \ref{sec:into}. From the WordNet subgraph, we compute $\lambda(T)$ as follows:

\begin{equation}
\lambda(T) = \frac {NodeSim(T) *ND(T)}{TC(T) *ED(T)}
\label{eq:lambda}
\end{equation}

\begin{itemize}
\item \textbf{NodeSim(T)}: This corresponds to the \textbf{average similarity between nodes representing the topics in the subgraph}. This is obtained by calculating the average cosine similarity between the corresponding node embedding, obtained via the \textit{TransE} multi-relational embedding learning technique \cite{bordes2013translating}. TransE operation on a graph results in low-dimensional embeddings of the nodes that capture its relationship with the other nodes in the graph in a distributional manner. In our setting, a higher similarity between two node embeddings indicate higher inter-relationship between them. We compute cosine similarity between the embeddings of each pair of nodes in the sub-graph and then average the similarity scores.

\item{\textbf{ND(T)}}: This represents the \textbf{average neighborhood degree} of the nodes in the subgraph. Intuitively, a higher average neighborhood degree indicates higher degree of connectedness amongst nodes, indicating higher topic relatedness.

\item{\textbf{ED(T)}}: This denotes the \textit{edge-density} of the graph\footnote{\url{https://en.wikipedia.org/wiki/Dense_graph}}. The notion behind using this measure is that if topics are distantly placed in the WordNet graph (a case of higher incoherence), the subgraph generated will be denser with more number of nodes and even more number of edges established through various WordNet relationships. So, higher $ED(T)$ should penalize $\lambda(T)$ and the overall coherence score.

\item{\textbf{TC(T)}}: We define this term as: 
$$ TC(T) = \frac{\# edges~in~the~subgraph}{\#edges~in~its~transitive~closure} $$
Since transitive closure of a graph indicates the node reachability, the term $TC$ reduces the reachability of nodes in the subgraph. For a graph like Wordnet where there are limited relations between nodes, subgraphs with non-ambiguous reachable paths are indicative of stronger topic-relatedness. Hence, a higher $TC$ score should penalize $\lambda(T)$ and the overall coherence score.
\end{itemize}

We provide an end-to-end algorithm for coherence score calculation in Algorithm \ref{alg:coherencescorecalculation}. 
\algsetup{indent=2em}
\newcommand{\coherencescorecalculation}{\ensuremath{\mbox{\sc Coherence Score Calculation }}}
\begin{algorithm}[t]
\caption{$\coherencescorecalculation$}
\label{alg:coherencescorecalculation}
\begin{algorithmic}[1]
\STATE \textit{function} \textbf{CalculateCoherenceScores} (Para $P$, WordNet graph $G$)
\newline

\STATE $T$ $\gets$ FindTopics ($P$)
\STATE $SG$ $\gets$ CreateSubGraph ($G$, $T$)
\STATE $E(T)$ $\gets$ calcEntropy ($T$, $P$)
\STATE $NodeSim(T)$ $\gets$ calcTransESimilarity ($SG$)
\STATE $ND(T)$ $\gets$ calcConnectivity($SG$)
\STATE $ED(T)$ $\gets$ calcEdgeDensity ($SG$)
\STATE $TC(T)$ $\gets$ calcTCScore ($SG$)
\newline
\STATE $\lambda(T) = \frac{NodeSim(T)\times ND(T)}{ED(T)\times TC(T)}$
\newline
\STATE $CS = \kappa \times E(T)\times\lambda(T)$
\newline
\RETURN $coherence\_score, CS$
\end{algorithmic}
\end{algorithm}

\section{Experimental Setup}
\label{sec:setup}
\subsection{Datasets}
We carry out our experiments on two sets of data as described below:
\subsubsection{Human Essay Data}
We take the Kaggle data released by the Hewlett Foundation for the task of Automated Essay Grading \cite{kaggledata}. The dataset consists of eight essay-sets corresponding to two types of essay prompts. We consider the persuasive/ narrative/ expository sets of essays (i.e., essay set id 1, 2, 7 and 8). Each essay is provided with scores of two or more human experts. We take the resolved expert scores for essay sets 1, 7 and 8; and mean of domain1 and domain2 scores for essay set 2 as gold labels. The expert scores indicate the overall goodness of the essay in terms of coherence, cohesion, organization, language-structure \textit{etc}. Though the overall grades are not exact labels for coherence, coherence plays an influential role while grading the essays. Hence, showing a positive correlation to these human graded essay scores can provide a validation for our coherence metric.\footnote{We refrain from creating a dataset with manually labeled coherence scores, due to the subjective nature of the labeling task}.
From the Kaggle data we extracted 5870 essays in total with varied number of sentences in each essay ranging from 1-84. More details on the dataset can be found in Table 2.

\subsubsection{Synthetic Data}
We created a synthetic data based on the essays provided by The Louvain Corpus of Native English Essays (LOCNESS) \cite{syntheticdata}. The corpus consists of argumentative and literary essays written by British and American university students. The essays are written on different topics ranging from \emph{computers}, \emph{biology}, \emph{British society } \textit{etc.}. We take the original essays and replace a fraction of the paragraph with sentences randomly chosen from essays on completely different topics. All the original paragraphs are labeled with a coherence score of 1.0. The coherence scores of the synthesized variants are determined by the degree of incoherency introduced. For example, if 20\% of the original paragraph is replaced with sentences from essays on a different topic, the coherence score is reduced by 20\%. If replaced sentences are extracted from essays on ``two'' different topics, the score is reduced further by another 20\%. Since many of the essays are extremely big, we sampled 81 essays (including variants) which had less than 1000 words. In the dataset, essays are labeled with coherence scores of C = [1.0,0.8,0.6,0.4] based on the above criteria, which is later treated as ground truth.

The state-of-the-art data-driven NLG systems that generate discourses often mix different topics (ref. abstractive summarization work discussed in the introduction). We try to mimic that by replacing portions of the coherent paragraphs with sentences from other paragraphs discussing unrelated topics. This is the rationale behind creating such a dataset for evaluation.

\subsection{Systems Details for $E(T)$ and $\lambda(T)$}
To cluster the GloVe word vectors we experimented with different values of K for K-means clustering algorithm. Finally, we chose K=1000 based on the values of average inter-cluster and intra-cluster distances. The conditional probability inference model discussed in Section 2.4 above was trained on a mixed domain corpus, \textit{i.e.}, UMBC\footnote{\url{http://ebiquity.umbc.edu/blogger/2013/05/01/umbc-webbase-corpus-of-3b-english-words/}} WebBase corpus of 3 billion English words.
The pre-computed transE embeddings trained on the Wordnet graph were obtained from \cite{bordes2013translating}. Our coherence scores were in the real-valued range [0.1-1000].

\subsection{Evaluation Metrics}
We obtained coherence scores $CS$ using Algorithm 2 for each of the four sets of human graded essays (referred to as $Set 1$, $Set 2$, $Set 7$ and $Set 8$ henceforth); and the synthetic data generated (referred to as $Synthetic$). We obtained Spearman's rank correlation coefficient between the gold labels and calculated coherence scores to see how they are correlated. Since, for most practical purposes, the relative ranking of paragraphs based on coherence may be more important than computing the absolute coherence scores, we chose the Spearman's rank correlation metric instead of Pearson's correlation. We also computed the \textit{rMSE} between the scaled coherence scores and gold labels for each of the five sets of data.  

\begin{table}[bht]
	\small
    \centering
	\begin{tabular}{l l l}
		\toprule
		\textbf{Dataset} & \textbf{\#Essays} & \textbf{Avg. \#Sents}\\
        \midrule
		Synthetic & 81 & 27.38 \\
		Set 1 & 1783 & 22.77 \\
		Set 2 & 1800 & 20.36 \\
		Set 7 & 1569 & 11.71\\
		Set 8 & 723 & 34.88\\
		\bottomrule
	\end{tabular}
	\label{tab:stats}
	\caption{Data statistics}
\end{table}
\begin{table}[bht]
	\small
    \centering
	\begin{tabular}{l l l}
		\toprule
		\textbf{Dataset} & \textbf{Correlation ($p$)} & \textbf{rMSE}\\
        \midrule
		Synthetic & 0.417 (1e-4) & 0.37 \\
		Set 1 & 0.502 (1.2e-114) & 0.63 \\
		Set 2 & 0.433 (4.3e-83) & 0.78 \\
		Set 7 & 0.411 (4.9e-65) & 0.72\\
		Set 8 & 0.283 (8.2e-15) & 0.46\\
		\bottomrule
	\end{tabular}
	\label{tab:main_results}
	\caption{Results for synthetic and human essay datasets. $Correlation \rightarrow$ Spearman's Rank Correlation Coefficient between measured and gold values of coherence with statistical significance test values $p$. 
	$rMSE \rightarrow$ Root Mean Squared Error between normalized measured and gold values of coherence. All correlation values are within 99\% confidence ($p<0.01$).}
\end{table}

\begin{table*}[t]
 \small
 \centering
	\begin{tabular}{l c c c c c}
		\toprule
		\textbf{Dataset} & $E(T)$ & $NodeSim(T)$ & $ED(T)$ & $TC(T)$ & $ND(T)$\\
        \midrule
		Synthetic & 0.255 & 0.107 & -0.084 & -0.015 & 0.042 \\ 
		Set 1 & 0.207 & -0.015 & -0.494 & -0.497 & 0.516 \\
		Set 2 & 0.205 & -0.115 & -0.384 & -0.438 & 0.496 \\ 
		Set 7 & 0.293 & 0.157 & -0.427 & -0.351 & 0.401 \\
		Set 8 & 0.08 & 0.142 & -0.385 & -0.28 & 0.356 \\
		\bottomrule
	\end{tabular}
	\label{tab:support_results}
	\caption{Spearman Rank Correlation between different components of the coherence metric and gold labels}
\end{table*}
\vspace{-3mm}
\section{Results and Discussion}
\label{sec:results}
\subsection{Correlation Analysis}
The results for the Spearman's rank correlation and rMSE between the coherence scores and gold labels are reported in Table 3. As it can be seen, in all cases our coherence scores obtain a positive correlation with the corresponding gold labels, suggesting that we are indeed modeling coherence, which plays an essential role in human essay grading. The low values of rMSE indicates that the predicted coherence scores are quite acceptable. To see the importance of each component in our coherence score, we calculated the component-wise Spearman's rank correlation coefficients w.r.t. the gold labels for each of the five datasets (Table 4). It can be seen that entropy and average neighborhood degree are positively correlated and TC score and edge density are negatively correlated as expected. The correlation with $NodeSim(T)$ is mostly positive, though for essay sets 2 and 7 we get slightly negative correlations. This might be due to the specific nature of prompts of these essay sets. Set 1 and Set 2 are persuasive essay prompts. Hence, the responses may contain complex relations between words which might not be captured by a WordNet graph which models very few specific kinds of relations. The comparatively high correlation of entropy component shows that it is an essential part of our coherence score formulation.
\vspace{-2mm}
\subsection{Topic Membership Visualization}
\begin{figure}[t]
\begin{center}
\includegraphics[width=6.75cm]{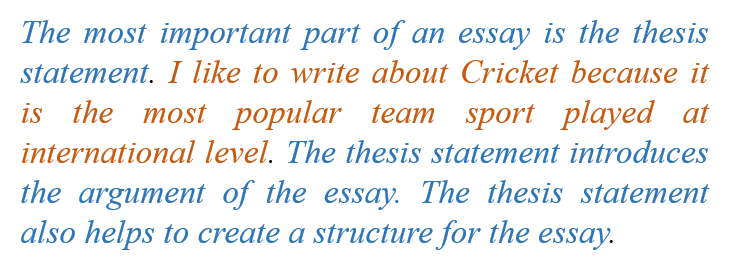}
\caption{Colour coded representation of topic membership for the incoherent paragraph.}
\label{fig:color_code}
\end{center}
\end{figure}

We ran our pipeline for Example 1 given in Table \ref{ref:example} (Coherent) and the example shown in Figure \ref{fig:color_code} (Incoherent). We got coherence score values of $17.29$ and $0.964$ respectively, for the two cases. In each case, out of all the topics obtained for the paragraph, we obtained the topic membership of each sentence. Interestingly for the second case, as shown in Figure \ref{fig:color_code}, out of the two topics obtained for the paragraph, the first, third and fourth sentences belong to one topic (coloured blue), and the second sentence belongs to the second topic (coloured red), indicating a significant drift in topic and hence incoherence.
\vspace{-2mm}
\subsection{Human Evaluation}
We conducted human evaluation on a random group of 27 essays (9 sets) from our synthetic dataset. The task was to rank each paragraph within a set (1 original and 3 perturbed variants) based on the coherence. We compared these rankings done by human subjects with the ranking produced by our system based on the values of coherence scores. We found an agreement of 79.3\% between the two rankings, which suggests that the rankings produced by our system are acceptable.
\vspace{-0.5mm}
\subsection{Comparison with Supervised Techniques}
Intrigued to see how our system performs as compared to supervised techniques for measuring local coherence, we tested our system on the \textit{Earthquakes} and \textit{Accidents} datasets released by \cite{barzilay2008modeling}. We obtained coherence scores for the essays given in the test set of the two datsets. Then the accuracy was measured by considering the fraction of test pairs ranked correctly based on the values of our coherence scores. If the relative difference between the coherence scores was less than a fixed tolerance, we marked them as positive. We got accuarcies of \textbf{77.3\%} and \textbf{71.5\%} on Earthquakes and Accidents datasets, respectively. These values are competitive to the reported accuracy figures of 81.4\% and 86.0\% (row Coreference-Syntax-Salience- in Table 5 of \cite{barzilay2008modeling}) on the two datasets. Considering the fact that our system is unsupervised and does not need additional complex meta-information like dependency parses and coreferences, syntax and saliency information etc., which the best supervised systems use today; our system offers significant advantages and is more generalizable compared to the popular supervised techniques.


\section{Related Work}
\label{sec:related}
The importance of discourse coherence analysis and measurement was identified long back by classical and computational linguists. Earlier works by \newcite{bamberg1983makes}, \newcite{ryan1984conceptions}, \newcite{mcculley1985writing} formalize coherence and properties of coherent discourses. 
There have been several works on automated essay grading \cite{attali2004automated}, modeling paragraph organizations \cite{persing2010modeling}, automated text scoring \cite{alikaniotis-yannakoudakis-rei:2016:P16-1}, and measuring coherence quality \cite{somasundaran2014lexical}. Our metric can certainly be used in these scenarios. 

In domains such as education, e-commerce, judicial and compliance many automatic scorers have been proposed over the last couple of decades. \cite{higgins2004evaluating} and \cite{miltsakaki2004evaluation} propose frameworks for measuring text coherence for essays collected by ETS. Since their data is not available publicly, a comparative study could not be carried out. \newcite{foltz1998measurement} propose a coherence model using latent semantic analysis. Using various textual and grammatical properties of the text, \newcite{crossley2011text} implemented a statistical regression based system for essay scoring and ranking. Recent works on evaluating the holistic scores of essays rely on deep learning based techniques \cite{alikaniotis-yannakoudakis-rei:2016:P16-1,taghipour2016neural}. However, relatively very little work has been done for individual aspects of the essay, such as organization \cite{persing2010modeling}, coherence and cohesion \cite{somasundaran2014lexical}. 

A significant amount of research has been carried out on modeling sentence ordering and local coherence in paragraphs. \newcite{barzilay2008modeling}, in a pioneering work, modeled local coherence in paragraphs (a comparison with them is provided in Section 4.4 above)
The rank-labels are predicted in a supervised setting with features extracted from the paragraphs based on an entity-grid model. Several works that addressed the problem of local coherence using the same (or similar) datasets are: (a) HMM based approach considering syntactic patterns by \newcite{louis2012coherence}, (b) Window Based Approach by \newcite{li2014model}, (c) Sequence-to-sequence based approach by \newcite{li2017neural}, and recurrent neural network based approach by \cite{logeswaran2016sentence}. These approaches, unlike ours, are supervised, and some of them require complicated meta-linguistic information to be extracted through Role labeling, Coreference resolution, dependency parses etc. thus requiring expensive additional resources.
\section{Conclusion and Future Work}
\label{sec:conclusion}
In this paper, we presented a metric for scoring paragraph topical coherence of natural language text on real valued scale. 
To measure topical congruency, our system first extracts a set of possible topics that emanate, as sentences in the paragraph unfold. Paragraph coherence is then measured as a product of two components capturing (a) paragraph-topic association, and (b) topic relatedness. Experiments on two datasets of human generated and automatically synthesized paragraphs reveal that the coherence scores produced by our system are positively correlated with the ground-truth. An additional human evaluation on a subset of synthetic dataset also proves the goodness of our measure, showing a strong agreement between coherence based ranking of paragraphs done by humans and our system. Our framework is \textbf{quite simple, unsupervised and highly modular}, making it possible to interpret, evaluate as well as plug-and-play individual components. Moreover, our framework offers the flexibility to \textbf{trivially extend it to other languages} with a Wordnet. Our future agenda includes introducing additional relevant components of coherence measurement into our formulation. We would also like to apply our metric to optimize NLG systems for abstractive summarization and descriptive QA. 
\vspace{-2mm}
\bibliography{emnlp2018}
\bibliographystyle{acl_natbib_nourl}
\end{document}